# Convolutional Neural Network-based Place Recognition

Zetao Chen, Obadiah Lam, Adam Jacobson and Michael Milford
ARC Centre of Excellence for Robotic Vision
Queensland University of Technology

## Abstract

Recently Convolutional Neural Networks (CNNs) have been shown to achieve state-of-the-art performance on various classification tasks. In this paper, we present for the first time a place recognition technique based on CNN models, by combining the powerful features learnt by CNNs with a spatial and sequential filter. Applying the system to a 70 km benchmark place recognition dataset we achieve a 75% increase in recall at 100% precision, significantly outperforming all previous state of the art techniques. We also conduct a comprehensive performance comparison of the utility of features from all 21 layers for place recognition, both for the benchmark dataset and for a second dataset with more significant viewpoint changes.

## 1 Introduction

Since their introduction in the early 1990s, Convolutional Neural Networks (CNNs) have been used to achieve excellent performance on a variety of tasks such as handwriting recognition and face detection. More recently, supervised deep convolutional neural networks have been shown to deliver high level performance on more challenging classification tasks [Krizhevsky, et al., 2012]. The key supporting factors behind these impressive results are their ability to learn tens of millions of parameters using large amounts of labelled data. Once trained in this way, CNNs have been shown to learn discriminative and human-interpretable feature representations [Zeiler and Fergus, 2013]. Most impressively, these approaches are capable of producing state of the art performance on tasks that the model was not explicitly trained for [Donahue, et al., 2013], including object recognition on the Caltech-101 dataset [Fei-Fei, et al., 2007], subcategory recognition on the Caltech-USCD birds dataset [Welinder, et al., 2010], scene recognition on the SUN-397 dataset [Xiao, et al., 2010] and object detection on the PASCAL VOC dataset [Girshick, et al., 2013]. This good generalization in performance on new tasks and datasets indicates that CNNs may provide a general and universal visual feature learning framework applicable to all tasks. Encouraged by these positive results, in this paper we develop a place recognition framework centered around features from pre-trained CNNs as illustrated in Figure 1.

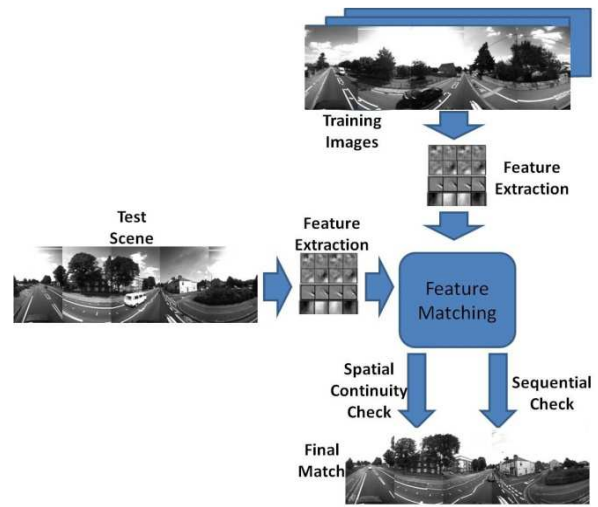

Figure 1 Schematic illustration of the deep learning-based place recognition system. Deep learning features are extracted from the test image and then matched to all training images. After the spatial and sequential continuity check, the final match is reported.

Place recognition can be considered as an image retrieval task which consists of determining a match between the current scene and a previously visited location. State-of-the-art visual SLAM algorithms such as FAB-MAP [Cummins and Newman, 2008] match the appearance of the current scene to a past place by converting the image into bag-of-words representations [Angeli, et al., 2008] built on local features such as SIFT or SURF. However, recent evidence [Krizhevsky, et al., 2012] suggests that features extracted from CNNs trained on very large datasets significantly outperform SIFT features on classification tasks. Donahue [Donahue, et al., 2013] shows that using mid-level features from CNN models trained on the ImageNet database can more efficiently remove dataset bias in some of the domain adaption studies than a bag of words approach.

The authors are with the School of Electrical Engineering and Computer Science at the Queensland University of Technology, Australia. http://roboticvision.org/. email: zetao.chen@student.qut.edu.au.
This work was supported by a funding from the Australian Research Council Centre of Excellence CE140100016 in Robotic Vision.

In this paper we investigate whether the advantages of deep learning in other recognition tasks carries over to place recognition. We present a deep learning-based place recognition algorithm that compares the response of feature layers from a CNN trained on ImageNet [Deng, et al., 2009] and methods for filtering the subsequent place recognition hypotheses. We conduct two experiments, one on a 70 km benchmark place recognition dataset, and one on a viewpoint varying dataset, providing both quantitative comparison to two state of the art place recognition algorithms and analysis of the utility of different layers within the network for viewpoint invariance.

The paper proceeds as follows. Section 2 provides an overview of feature-based place recognition techniques and convolutional neural networks. In Section 3 we describe the components of the deep learning-based place recognition system. The experiments are described in Section 4, with results presented in Section 5. Finally we conclude the paper in Section 6 and discuss ongoing and future work.

## 2 Related Work

In this section, we briefly review feature-based representations for place recognition and the use of convolutional neural networks for various visual classification tasks.

### 2.1 Vision Representation for Place Recognition

Visual sensors are increasingly becoming the predominant sensor modality for place recognition due to their low cost, low power requirements, small footprint and rich information content. There has been extensive research on how to best represent and match images of places.

Several authors have described approaches that apply global feature techniques to process incoming sensor information. In [Murillo and Kosecka., 2009], the authors propose a gist feature-based place recognition system using panoramic images for urban environments. Histograms of image gray values or texture is also a widely used feature in place recognition systems [Ulrich and Nourbakhsh, 2000, Blaer and Allen, 2002] due to its compact representation rotation invariance. However, global features are computed from the entire image, rendering them unsuitable to effects such as partial occlusion, lighting change or perspective transformation [Deselaers, et al., 2008].

Local features are less sensitive to these external factors and have been widely used in appearance-based loop closure detection, SIFT [Lowe, 1999] and SURF [Herbert Bay, et al., 2008] being two widespread examples. State-of-the-art SLAM systems such as FAB-MAP [Cummins and Newman, 2008] further represent appearance data using sets of local features, converting images into "bag-of-words", which enables efficient retrieval. Other feature-less representations have also been proposed. SeqSLAM [Milford and Wyeth, 2012] directly uses pixel values to match image sequences and perform place recognition across extreme perceptual changes. However, it is rapidly becoming apparent in other recognition tasks that hand-crafted features are being outperformed by learnt features, prompting the question of whether we can learn better features automatically?

### 2.2 Convolutional Neural Networks

Convolutional neural networks are multi-layer supervised networks which can learn features automatically from datasets (Figure 3). For the last few years, CNNs have achieved state-of-the-art performance in almost all important classification tasks [Krizhevsky, et al., 2012, Donahue, et al., 2013, Sharif Razavian, et al., 2014]. Their primary disadvantage is that they require very large amounts of training data. However, recent studies have shown that state of the art performance can be achieved with networks trained using "generic" data, raising the possibility of developing a place recognition system based on features learnt from datasets with a classification focus. A similar approach has already achieved excellent performance on various visual tasks, such as object recognition [Fei-Fei, et al., 2007]; subcategory recognition [Welinder, et al., 2010]; scene recognition [Xiao, et al., 2010] and detection [Girshick, et al., 2013].

One research area separate but relevant to the place recognition problem is the task of image retrieval where a query image is present to a database to search for those images containing the same objects or scenes. In [Babenko, et al., 2014], mid-level features from CNNs are evaluated for the image retrieval application and achieve performance comparable to others using state-of-the-art features. Interestingly, the best performance is obtained using mid-network features rather than those learnt at the final layers.

Place recognition is essentially a task of image similarity matching. In [Fischer, et al., 2014], features from various layers of CNNs are evaluated and compared with SIFT descriptors on a descriptor matching benchmark. The benchmark results demonstrate that deep features from different layers of CNNs consistently perform better than SIFT on descriptor matching; indicating that SIFT or SURF may not be the preferred descriptors for matching tasks anymore. Our paper is thus inspired by the excellent performance of CNNs on image classification and the evidence of their feasibility in feature matching.

## 3 Approach and Methodology

In this section we describe the two key components of our approach: feature extraction and spatio-temporal filtering of place match hypotheses output by comparison of feature responses. The schematic illustration of the method procedure is shown in Figure 2:

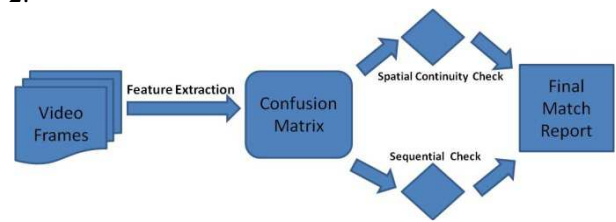

Figure 2 The place recognition process. A confusion matrix is constructed from features, resulting in place match hypotheses which are then filtered to produce finalized matches.

### 3.1 Feature Extractor

We use a pretrained network called Overfeat [Sermanet, et al., 2013] which was originally proposed for the ImageNet Large Scale Visual Recognition Challenge 2013 (ILSVRC2013). The Overfeat network is trained on the ImageNet 2012 dataset, which consists of 1.2 million images and 1000 classes. The network comprises five convolution stages and three fully connected stages (Figure 3). Each of the bottom two convolution stages consists of a convolution layer, a max pooling layer and a rectification (ReLU) non-linearity layer. The third and fourth stages consist of a convolution layer, a zero-padding layer and a ReLU non-linear layer. The fifth stage contains a convolution layer, a zero-padding layer, a ReLU layer and a max-pooling layer. Finally, the sixth and seventh stages contain one fully-connected layer and one ReLU layer, while the eighth contains only the fully-connected output layer. In total, there are 21 layers.

When an image I is input into the network, it produces a sequence of layered activations. We use $L_k(I), k = 1, ..., 21$ to denote the corresponding output of the $k^{th}$ layer given input image I. Each of these vectors is a deep learnt representation of the image I; place recognition is performed by comparing these feature vector responses to different images. The network is capable of processing images of any size equal to or greater than $231 \times 231$ pixels, consequently all experiments described here used images resized to $256 \times 256$ pixels.

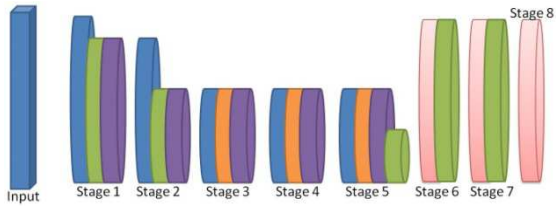

Figure 3 Architecture of the Overfeat network.

### 3.2 Confusion Matrix

For each layer output $L_k(I), k = 1, ..., 21$, we generate a corresponding confusion matrix $M_k, k = 1, ..., 21$ from the whole dataset with $R$ training images and $T$ testing images (Figure 4). Each element $M_k(i, j)$ represents the Euclidean distance between the feature vector responses to the $i^{th}$ training image and the $j^{th}$ testing image:

$$M_k(i, j) = d(L_k(I_i), L_k(I_j))$$
$$i = 1, ..., R, j = 1, ..., T \quad (1)$$

Each column $j$ stores the mean feature vector difference between the $j^{th}$ testing image and all training images.

To find the strongest place match hypothesis, each column is searched for the element with the lowest feature vector difference.

$$M_k(j) = \arg\min_i M_k(i, j), \forall i, j = 1, ..., T \quad (2)$$

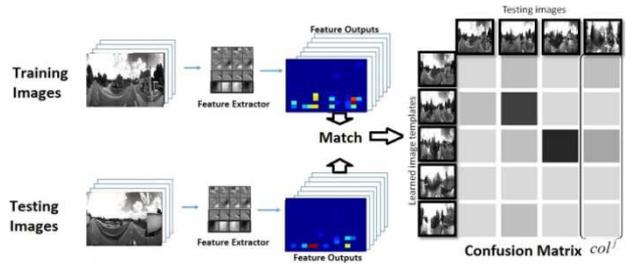

Figure 4 The procedure for generating a confusion matrix. Features are extracted by Overfeat from each testing image and then matched to features from all the training images. The matrix element $M(i,j)$ represents the Euclidean distance between the $i^{th}$ training image and the $j^{th}$ testing image.

### 3.3 Spatial Continuity

We apply two continuity filters to the place match hypotheses extracted from the confusion matrix. The first, a spatial continuity check, enforces that consecutive first-ranked place match hypotheses must occur in close indices in the confusion matrix, providing a constraint that does not require a specific motion model. More specifically, the plausible measurement $P_k(j)$ of each place match hypothesis is evaluated as follows:

$$P_k(j) = 1$$
$$if \; |M_k(u-1) - M_k(u)| \le \varepsilon, \forall u \in [j-d, j] \quad (3)$$
$$j = d, ..., T$$

where $\varepsilon$ is the threshold for consecutive first-ranked match difference, $d$ determines how far back in time the evaluation goes and $j$ is the current testing image. A positive match is reported only when $P(j) = 1$.

Figure 5 provides an illustration of the spatial continuity check in action. This constraint reduces but does not eliminate all false positives; consequently we implement a secondary, sequential filter step that implements an actual motion model, and is described in the next section.

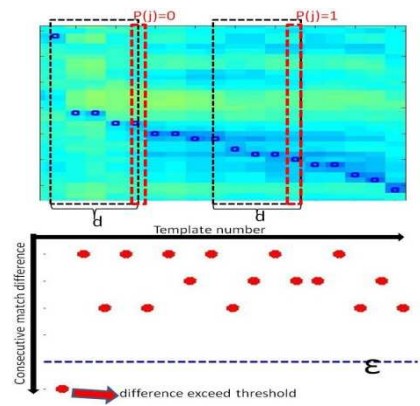

Figure 5: Illustration of spatial continuity constrain. Blue squares indicate where $M(j)$ is in each column. The left red dot square represents a non-plausible match because one of the consecutive match difference within the black dot square (evaluation window) exceeds the threshold ε. The right red dot square indicates a plausible match.

### 3.4 Sequential Filter

The sequential filter is a more sophisticated implementation of the crude motion filter in SeqSLAM.

Rather than searching for all coherent diagonal sequences of strong matching hypotheses, we only fit linear polynomial models to the top matches in each local sequence $S_j = \{M(j-d), M(j-d+1), ..., M(j)\}$ by using:

$$y = f(x) = \alpha_j x + \beta_j \quad (4)$$

where $d$ is the sequence length used in Section 3.3, $j$ is the current frame and $\alpha_j$ describes the slope of the linear model in sequence $S_j$ which represents the velocity ratio between the second and first traverse. As shown in Figure 6, the place match hypotheses $F(j)$ comprising a sequence are accepted if the velocity ratio is within a certain bound $\varphi$ around a reference velocity $\sigma$. The parameter $\varphi$ is swept over a range of values to generate the precision-recall curves shown in Section 5.1. We also note here that if an odometry source is available, this sequence search could be significantly simplified in a manner similar to the SMART approach [Pepperell, et al., 2013].

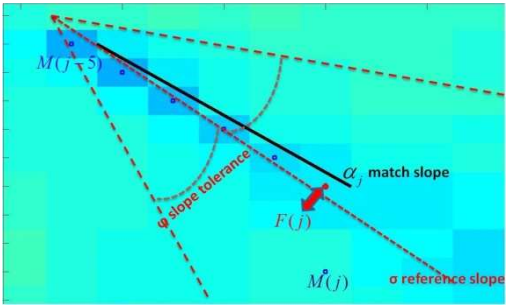

Figure 6 Describe how a final place match hypothesis $F(j)$ is estimated by fitting a linear model to a local sequence $S(j) = \{M(j-5), ..., M(j)\}$. The slope of the linear model $\alpha_j$ is within a bound $\varphi$ around a reference velocity $\sigma$ and therefore, the final match $F(j)$ generated by the linear model is considered as a plausible match.

## 4 Experimental Setup

In this section, we describe the datasets used, ground truth measures, and parameter values.

### 4.1 Datasets

Details of the two datasets used are summarized in Table 1. Each dataset consists of two traverses along the same route, with the first traverse used for training and the second traverse used for testing. For both environments, full resolution images were first converted to gray-scale and then histogram normalized to reduce the effect of illumination variations. Images were then resized to 256 × 256 pixels before being input into the CNNs.

The Eynsham dataset is a large 70 km road-based dataset (2 × 35 km traverses) used in the [Cummins and Newman, 2009] FAB-MAP and SeqSLAM studies. Panoramic images were captured at 7 meter intervals using a Ladybug 2 camera. The QUT dataset was collected using a hand-held camera walking around the Queensland University of Technology campus, with a viewpoint shift of up to 5 metres lateral camera movement between the first and second traverses.

### 4.2 Ground Truth

For the Eynsham dataset, we used the 40 metre tolerance GPS-derived ground truth provided with the Eynsham dataset, consistent with the tolerance used in the original FAB-MAP study [Cummins and Newman, 2008] and SeqSLAM study [Milford and Wyeth, 2012]. For the QUT dataset, ground truth was obtained by manually parsing each frame and building frame correspondence. We use a tolerance of 2 frames, corresponding to approximately 3.8 meters.

| Dataset Name | Total Distance | Total Number of Frames | Distance between frames |
|---|---|---|---|
| Eynsham | 70 km | 9575 | 6.7 m (median) |
| QUT | 380 m | 200 | 1.9 |

Table 1 Dataset Description…

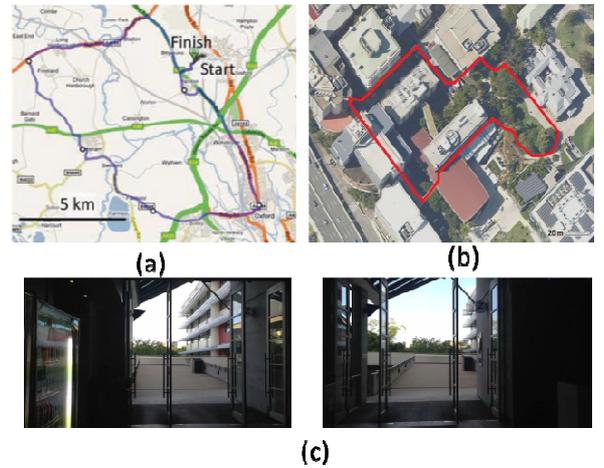

Figure 7: Aerial overhead images showing the dataset route for the (a) 70 km Eynsham and (b) 300m QUT dataset. Imagery @2014 Google, Map data @2014 Google. (c) Illustration of the moderate viewpoint variation in the QUT dataset.

### 4.3 Parameter Values

In Table 2, we provide the values of the critical parameters used in the experiment.

| Parameter | Value | Description |
|---|---|---|
| $\varepsilon$ | 3 | Threshold for local spatial constrain |
| d | 5 | Length of local sequence for polynomial model fit |
| $\sigma$ | $\pi/4$ | Prior knowledge about the matching slope |

Table 2 Parameter values.

## 5 Results

In this section we present two sets of results on the Eynsham and QUT datasets; firstly, a comparison of performance between our proposed approach with both FAB-MAP 2.0 and SeqSLAM on the benchmark Eynsham dataset, as well as an evaluation of the performance of each feature layer for viewpoint invariant place recognition. We also provide compute performance statistics and discuss the feasibility of a real-time implementation.

## 5.1 Precision-Recall curves

This section presents precision-recall curves on the Eynsham dataset for the deep learning-based place recognition algorithms using features from all layers of the CNNs with comparison to SeqSLAM and FAB-MAP. Each precision-recall curve was generated by performing a parameter sweep on slope tolerance $\varphi$ discussed in Section 3.4.

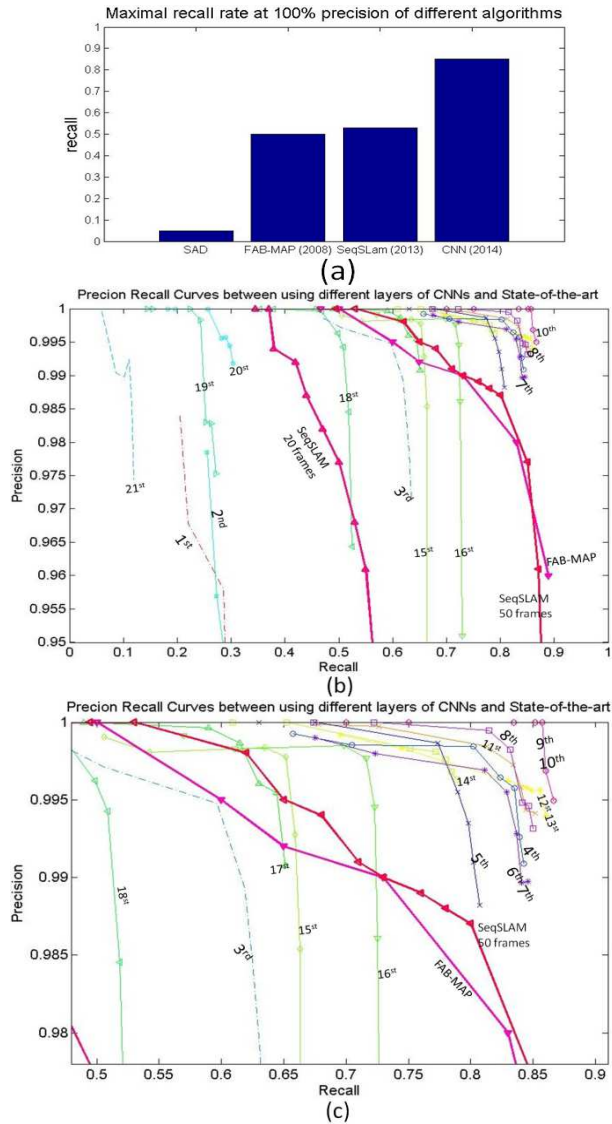

Figure 8 Precision-recall curves on the Eynsham dataset. (a) Maximal recall rates at 100% precision using different algorithms; (b) Precision-recall curves of deep learning method and state-of-the-art method, 19[th] indicates the result using features from the 19[th] layer of the network; (c) Zoom in on a particular section in (b).

Figure 8(a) demonstrates the maximum recall rates at 100% precision achieved using the best performing feature layers and SeqSLAM and FAB-MAP. The maximum recall rate that can be achieved by the deep learning-based approach is 85.7% compared to the approximately 51% recall rate achieved by SeqSLAM. Also noteworthy is that this result is achieved using a filter analogous to SeqSLAM operating with a sequence length of 5, rather than the sequence length of 50 with which the 51% recall performance is achieved.

Figures 8(b) and 8(c) present the precision-recall performance curves for all layers. Clearly the middle network layers provide the best performance, a result consistent the image retrieval experiments of [Babenko, et al., 2014] which suggest that the middle network layers provide a more general feature description while the top layers are overtrained for the ImageNet task.

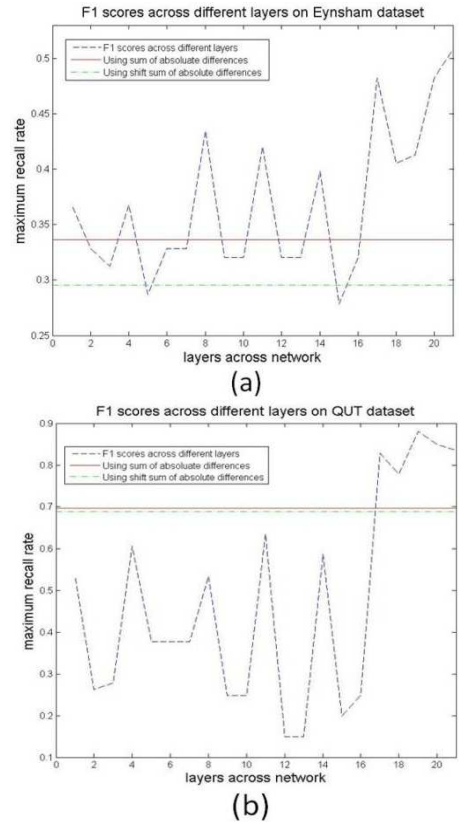

Figure 9 F1 scores on the QUT dataset and Eynsham subset across layers. (a) Results on the Campus dataset; (b) Results on the Eynsham subset;

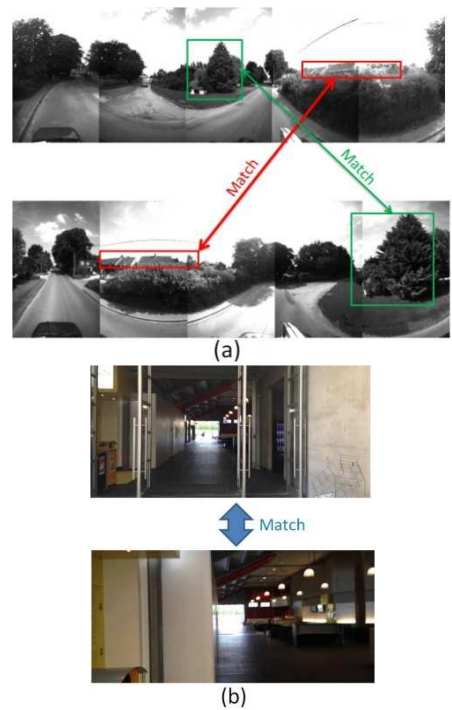

Figure 10 Sample place matches from (a) reverse trajectories on Eynsham dataset (b) and lateral viewpoint change on the QUT dataset.

## 5.2 Viewpoint Invariance

We evaluate the viewpoint invariance of different network layers using a custom QUT dataset specifically gathered with lateral camera variance, and by selecting a subsection of the Eynsham dataset where the car is travelling in the reverse direction on the opposite side of the road.

Figure 9 shows the F1 scores achieved by each layer of the CNN. There is a clear trend for increasing viewpoint invariance in later network layers. The red lines represent a performance baseline calculated by using a Sum of Absolute Differences (SAD) image comparison. The green dotted lines show the performance of SAD with offset matching, a measure frequently used to increase the viewpoint invariance of such as technique [Milford, 2013].

## 5.3 Visualization of Confusion Matrix

To provide a qualitative illustration of the superiority of using deep learnt features over simple techniques like SAD, we present a comparison between a subsection of the confusion matrix from the 10$^{th}$ layer generated using each approach (Figure 11).

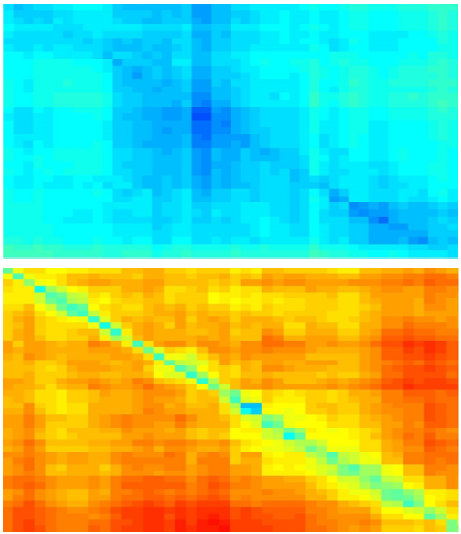

Figure 11 Comparison of confusion matrix from the 10$^{th}$ layer using sum of absolute difference (top) and deep learning features (bottom). A more clear diagonal pattern can be observed in the bottom figure.

## 5.4 Feasibility of Real-time Operation

Deep learning approaches are notoriously computationally-intensive so an examination of real-time capability is particular necessary. The experiments in this paper used the Overfeat network, and feature extraction ran at significantly slower than real-time on a single PC. However, we have recently re-implemented the system using another convolutional architecture dubbed Caffe [Jia, et al., 2014], and initial studies suggest that recognition performance is near-identical while being many orders of magnitude faster. Consequently, here we present a calculation of the computation required for extracting features and feature matching using Caffe.

*CNN Feature Extraction*

Using a standard CPU, 10 images can be processed per second.

*Feature Matching*

The largest feature vector is from the 10$^{th}$ layer which contains 64899 uint16 values. Comparing it to all 4789 training images will require:

$64899 \times 2 \times 4789 = 6.22 \times 10^8$ comparison/s

Based on this, a standard CPU should be able to process about 2.5 frames per second.

## 6 Discussion and Future Work

The results in this paper demonstrate that the task of place recognition can benefit immensely from incorporating features learnt using CNNs. In particular, performance using even a relatively simple framework around these features results in performance significantly better than the current state of the art algorithms. Furthermore it is interesting to note that different layers appear to be suitable for different aspects of the place recognition task – the middle layers being optimal for recognition on relatively static, similar viewpoint datasets, while later layers appear to perform better when viewpoint variance becomes significant. In this section we discuss some of the new opportunities and challenges that exist in this field.

### 6.1 Network Adaption Training

Datasets are inherently biased in computer vision [Torralba and Efros, 2011]. In [Hoffman, et al., 2013], the researchers demonstrated that a supervised deep CNN model trained on large amounts of labelled data reduces, but does not remove, data bias. The network we use in this paper is trained for a different classification task; therefore, although it demonstrates impressive generalization performance to a different recognition task, a major question remains unanswered; whether performance can be further improved by training a network from scratch with place recognition datasets. One potential problem with such an approach is the relative sparsity of very large place recognition datasets in comparison with the millions of frames found in the Imagenet database. One option will be to keep all the parameters from the pre-trained model and then add a final domain specific classification layer trained for each particular new dataset. This approach has been adopted in some domain adaption work from CNNs for object recognition [Hoffman, et al., 2013, Oquab, et al., 2013, Rodner, et al., 2013].

### 6.2 Automatic Layer Selection

Currently there is no mechanism for automatically select the best layer for the specific place recognition task. Future work will investigate automating selection of the best performing layers, most immediately by introducing a performance measurement for each layer during the training process.

### 6.3 Feature Ranking for Deep Learning Features

In this paper, we use a simple Euclidean distance metric to compare the similarity of feature responses, an approach that implicitly assumes that each feature contributes equally to place recognition performance. This assumption is likely unreasonable because feature weights are normally dataset-dependent. A feature which contributes strongly in one dataset may have little classification power for another dataset. In future work, we plan to train a dataset-dependent feature ranking algorithm for each new task to automatically weight the contributions of different features.